# Localized Multiple Kernel Learning for Anomaly Detection: One-class Classification

Chandan Gautam[a,∗], Ramesh Balaji[a], Sudharsan K.[a], Aruna Tiwari[a], Kapil Ahuja[a]

[a]*Indian Institute of Technology Indore, Indore, Simrol, India*

**Abstract**

Multi-kernel learning has been well explored in the recent past and has exhibited promising outcomes for multi-class classification and regression tasks. In this paper, we present a multiple kernel learning approach for the One-class Classification (OCC) task and employ it for anomaly detection. Recently, the basic multi-kernel approach has been proposed to solve the OCC problem, which is simply a convex combination of different kernels with equal weights. This paper proposes a **L**ocalized **M**ultiple **K**ernel learning approach for **A**nomaly **D**etection (*LMKAD*) using OCC, where the weight for each kernel is assigned locally. Proposed *LMKAD* approach adapts the weight for each kernel using a gating function. The parameters of the gating function and one-class classifier are optimized simultaneously through a two-step optimization process. We present the empirical results of the performance of *LMKAD* on 25 benchmark datasets from various disciplines. This performance is evaluated against existing **M**ulti **K**ernel **A**nomaly **D**etection (*MKAD*) algorithm, and four other existing kernel-based one-class classifiers to showcase the credibility of our approach. *LMKAD* achieves significantly better Gmean scores while using a lesser number of support vectors compared to *MKAD*. Friedman test is also performed to verify the statistical significance of the results claimed in this paper.

## 1. Introduction

The Anomaly Detection is a problem of finding instances of the input data that do not conform to the general pattern or behavior exhibited by the majority of the data points. This problem has been well explored and addressed in the past using One-class Classification [1, 2, 3, 4, 5, 6, 7, 8]. The One-class Classification (OCC) problem is unlike the conventional binary and multi-class classification problems in that in OCC, one has data about only one of the many

∗Corresponding author
*Email addresses:* chandangautam31@gmail.com (Chandan Gautam), ee140002025@iiti.ac.in (Ramesh Balaji),
cse140001014@iiti.ac.in (Sudharsan K.), artiwari@iiti.ac.in (Aruna Tiwari), kahuja@iiti.ac.in (Kapil Ahuja)

classes that could constitute the input space. Outliers, on the other hand, could belong to any class or even be isolated anomalies. Various models to handle OCC problems have been developed in the past [1]. Out of these, SVM based methods have gotten more attention from researchers due to their efficiency, kernel learning ability, and generalization capability. In SVM based methods, the OCC problem is redefined as the task of devising a boundary around the given data of target class points, such that most of the target class points lie within the defined boundary. Two types of SVM based methods have been developed viz., Support Vector Data Description ($SVDD$) [9] and One-class SVM ($OCSVM$) [10]. Tax and Duin [9] developed SVDD by finding a hyper-sphere of minimum radius around the target class data such that it encloses almost all points in the target class data set. Scholkopf et al. [10] extend the idea of $SVM$ for binary classification [11] to the domain of anomaly detection by proposing $OCSVM$. They construct a hyper-plane such that it separates all the data points from the origin and the hyper-plane's distance from the origin is maximum. We choose Scholkopf's $OCSVM$ as the classifier for our developments because of its robust performance as reported by other researchers [12]. It also guaranteed convergence and the flexibility of kernel methods in general [8, 13].

The general idea of kernel-based methods such as $SVM$ is to project the input space to a higher dimension where they become linearly separable. Kernel-based methods have received considerable attention over the last decade due to their success in classification problems. An advance in kernel-based methods for classification problems is to use many different kernels or different parameterization of kernels instead of a single fixed kernel [14] to get better performance. Using multiple kernels gives two advantages. Firstly, this provides flexibility to select for an optimal kernel or parameterizations of kernels from a larger set of kernels, thus reducing bias due to kernel selection and at the same time allowing for a more automated approach [15, 16, 17]. Secondly, multiple kernels are also reflective of the need to combine knowledge from different data sources (such as images and sound in video data). Thus they accommodate the different notions of similarity in the different features of the input space. A further advance in multiple kernel learning is to localize the kernel selection process for various task like, binary classification [18], image recognition problems [19], regression [20], and clustering [21]. Gonen et al. [18, 19, 20] achieve this localization by using a gating function which helps in selection of the appropriate kernel locally. Further this concept is explored by various researchers [22, 23, 24, 25].

There have been few attempts to transfer the idea of multiple kernel learning to the domain of One-class Classification and Anomaly Detection. Das et al. [26] propose a simple weighted sum of two kernels, each of which describes the discrete and continuous streams in aviation data, respectively. This method takes advantage of multiple kernel learning to incorporate the different notions of similarity in the two streams for the task of detecting anomalies in a heterogeneous system. Although, the method takes advantage of the ability of multi-kernel learning to combine knowledge from different data sources and exhibited better performance compared to single kernel learning. However, kernels are combined using only fixed combination rule, i.e., assign equal weight to each kernel over whole input space. In this paper, the proposed method provides more flexibility to the existing *MKAD* algorithm by taking advantage of local information presents among data and extend it to a localized formulation of multiple kernel-based *OCSVM* for anomaly detection, which is named as *LMKAD*. The intuition behind this work is same as discussed above and also by Jacobs et al. [27] and Gonen et al. [18] for multi-class classification. In this paper, we are extending this concept for anomaly detection using One-class Classification.

**The main contributions of this paper are as follows:**

1. This paper presents an optimization problem for data-driven anomaly detection in which a convex combination of kernels is used, with weights assigned locally. This optimization problem is analogous to the conventional *OCSVM* and can be solved similarly.

2. Our algorithm achieves significantly better Gmean scores compared to conventional *OCSVM* and *MKAD*, and at the same time uses a lesser number of support vectors compared to *MKAD*. For demonstrating the credibility of our algorithm, we perform extensive testing using five-fold Cross-validation (CV) over 5 runs on 13 small and 12 medium sized benchmark datasets.

3. *LMKAD* performance is compared with 5 state-of-the-art kernel-based methods available in the literature. Finally, a Friedman test [28] is conducted to verify the statistical significance the experimental outcomes of *LMKAD* classifier and it rejects the null hypothesis with 95% confidence level.

The rest of the paper is organized as follows. Sections 2 and 3 describe the *OCSVM* and Multiple Kernel Anomaly Detection (*MKAD*) algorithms, respectively. In Section 4, we propose *OCSVM* based Localized Multiple Kernel Anomaly Detection (*LMKAD*). Section 5 describes the experimental setup and evaluates the proposed (*LMKAD*) and existing One-class Classifiers (*OCSVM* and *MKAD*) against 25 benchmark datasets. The paper concludes in Section 6.

**2. One-Class SVM**

One-class SVM was proposed by Scholkopf et al. [10] for extending the utility offered by *SVM* to One-class classification. Given a set of training vectors $x_i \in \mathbb{R}^n, i = 1, \ldots, N$, where, all training vectors belong to the same class. *OCSVM* constructs a hyperplane that basically separates all the target class data points from the origin and maximizes the distance of this hyperplane from the origin. This is done by solving the following optimization problem.

$$\min_{\omega,\xi,\rho} \quad \frac{1}{2}\omega^T\omega - \rho + \frac{1}{\nu N}\sum_{i=1}^{N}\xi_i$$
$$\text{s.t.} \quad \omega^T\phi(x_i) \geq \rho - \xi_i \quad i = 0,\ldots,N,$$
$$\xi_i \geq 0, \quad i = 0,\ldots,N \tag{1}$$

The dual of which can be written as

$$\min_{\alpha} \quad \frac{1}{2}\alpha^T Q\alpha$$
$$\text{s.t.} \quad 0 \leq \alpha_i \leq \frac{1}{\nu N} \quad i = 0,\ldots,N,$$
$$\sum_{i}^{N}\alpha_i = 1 \tag{2}$$

where $Q_{ij} = K(x_i, x_j) = \phi(x_i)^T\phi(x_j)$

In the above two equations, $w$ is the weight coefficients, $\phi(.)$ is the mapping in the feature space, $K$ is the kernel matrix, $\alpha_i$ is the Lagrange multiplier, $N$ is the total number of training samples provided, $\nu$ is a parameter that lets the user define the fraction of target class points rejected, and $\rho$ is the bias term. This results in a binary function which returns $+1$ or $-1$ for target class and outliers, respectively, and is called the decision function. The decision function $f(x)$ thus obtained is as follows:

$$f(x) = sign(\sum_{i=1}^{N}\alpha_i K(x_i, x) - \rho) \tag{3}$$

Based on the formulation OCSVM, Das et al. [26] proposed anomaly detection for more than one kernel, which is described in the next section.

## 3. Multiple Kernel Anomaly Detection

Das et al. [26] proposed *MKAD* to detect anomalies in aviation data. Aviation data consists of features that can be grouped into two categories - (*i*) Real-valued data such as flight velocity, altitude, flap angle, etc and (*ii*) Binary valued data such as cockpit switch positions. Single-kernel *OCS VM* cannot capture the different notions of similarity in the Real and Binary valued data. Instead, a composite kernel $K$ is used. This composite kernel can be any valid convex combination of individual kernels.

This is the method used by Das et al. [26]. Here the composite kernel is a simple weighted sum of the individual kernels computed over all or a subset of the features, i.e.,

$$K(x_i, x_j) = \sum_{m=1}^{p} \eta_m k_m(x_i, x_j),\qquad(4)$$

where $\eta_m \geq 0$ and $\sum_{m=1}^{p} \eta_m = 1$. Here $k_m(x_i, x_j)$ represents the $m^{th}$ kernel computed for data points $x_i$ and $x_j$, and $\eta_m$ denotes assigned weight to individual kernels. The dual of this optimization problem is similar to that of $OCSVM$, with the kernel replaced by the composite kernel.

Note that here, the advantage of the multiple kernel learning approach is to incorporate knowledge of the differing notions of similarity in the decision process. Thus, we are able to achieve an improvement in detecting anomalies in a system that involves various data sources. A fixed combination rule (like a weighted summation or product) assigns the same weight to a kernel which remains fixed over the entire input space. However, this does not take into account the underlying localities in the data. Assigning different weights to a kernel in a data-dependent way may lead to a further improvement in detecting the anomalies. We explore this possibility in the next section.

## 4. Localized Multiple Kernel Anomaly Detection

In this section, we propose Localized Multiple Kernel Anomaly Detection ($LMKAD$). By assigning weights in a data-dependent way we intend to give more weights to kernel functions which best match the underlying locality of the data in different regions of the input space. We modify the decision function in the previous sections to the following:

$$f(x) = \sum_{m=1}^{p} \eta_m(x)\langle\omega_m, \phi_m(x)\rangle - \rho \qquad(5)$$

where $\eta_m(x)$ is the weight corresponding to each kernel and is assigned by the gating function. The value of $\eta_m(x)$ is a function of the input $x$ and is defined by the parameters of the gating function. These parameters are in turn learned from the data during optimization as shown later in this section. We rewrite the conventional $OCSVM$ optimization problem with our new decision function to get the following primal optimization problem

$$\min_{\omega_m, \eta_m(x), \xi, \rho} \quad \frac{1}{2}\sum_{m=1}^{p} \omega_m^T \omega_m - \rho + \frac{1}{\nu N}\sum_{i=1}^{N} \xi_i$$

$$\text{s.t.} \quad \sum_{m=1}^{p} \eta_m(x)\langle \omega_m, \phi_m(x)\rangle \geq \rho - \xi_i \quad \forall i, \qquad (6)$$

$$\xi_i \geq 0 \quad \forall i$$

where $\nu$ is the rate of rejection, $N$ is the total number of training samples, and $\xi_i$ are the slack variables as usual. We now need to solve this optimization problem for the above parameters. However, we do not solve this optimization problem directly, but use a two-step alternate optimization scheme inspired by Rakotomamonjy et al. [29] and Gonen et al. [18] to find the values of the parameters of the gating function ($\eta_m(x)$) and the parameters of the decision function.

Before starting the optimization procedure, we initialize the value of $\eta_m(x)$. Then, **in the first step** of the procedure we treat $\eta_m(x)$ as a constant and solve the optimization problem (6) for $\omega, \xi$ and $\rho$. Note that if we treat $\eta_m(x)$ as a constant, this step is essentially the same as solving conventional *OCSVM* under certain conditions as we will explain shortly. Solving the conventional *OCSVM* returns the optimal value of the Objective function and the Lagrange multipliers. **In the second step** we update the value of the parameters of $\eta_m(x)$ using gradient descent on the Objective function. The updated parameters define a new $\eta_m(x)$ which is used for the next iteration. The above two steps are repeated until convergence. From the Objective function in (6) and the constraints, for fixed $\eta_m(x)$ the Lagrangian of the primal problem is written as:

$$L_D = \frac{1}{2}\sum_{m=1}^{p} \omega_m^T \omega_m + \sum_{i=1}^{N}\left(\frac{1}{\nu N} - \beta_i - \alpha_i\right)\xi_i - \rho - \sum_{i=1}^{N} \alpha_i \left(\sum_{m=1}^{p} \eta_m(x_i)\langle \omega_m, \phi_m(x_i)\rangle - \rho\right) \qquad (7)$$

and taking the derivatives of the Lagrangian $L_D$ with respect to the variables in (6) gives :

$$\frac{\partial L_D}{\partial \omega_m} \Rightarrow \omega_m = \sum_{i=1}^{N} \alpha_i \eta_m(x_i)\phi_m(x_i) \quad \forall m \qquad (8)$$

$$\frac{\partial L_D}{\partial \rho} \Rightarrow \sum_{i=1}^{N} \alpha_i = 1 \qquad (9)$$

$$\frac{\partial L_D}{\partial \xi_i} \Rightarrow \frac{1}{\nu N} = \beta_i + \alpha_i \qquad (10)$$

Substituting, (8), (9), and (10) into (7), we obtain the dual problem,

$$\max_{\alpha} \quad J(\eta) = -\frac{1}{2}\alpha^T Q \alpha$$

$$\text{s.t.} \quad 0 \leq \alpha_i \leq 1 \quad i = 0, \ldots, l, \tag{11}$$

$$\sum_i^N \alpha_i = 1$$

where $Q_{ij} = K_\eta(x_i, x_j)$. Here, the kernel matrix is defined as:

$$K_\eta(x_i, x_j) = \sum_{m=1}^{p} \eta_m(x_i)\langle \phi_m(x_i), \phi_m(x_j)\rangle \eta_m(x_j) \tag{12}$$

The objective function of the dual is termed as a function of $J(\eta)$. The dual formulation is exactly the same as the conventional *OCSVM* formulation with kernel function $K_\eta(x_i, x_j)$. Multiplying the kernel matrix with a non-negative value will still give a positive definite matrix [30]. Note that the locally combined kernel function will, therefore, satisfy the Mercer's condition if the gating function is non-negative for both input instances. This can be easily ensured by picking a non-negative $\eta_m(x)$. In order to assign the weight to the different kernels based on the training data, a gating function is used.

In above formulations of *LMKAD*, objective value of the dual formulation (11) is equal to the objective value of the primal (6). By using the dual formulation in (11), gating function is trained and $\eta_m(x)$ is computed. For training, the gradient of the objective function of the dual (i.e. $J(\eta)$) is computed with respect to the parameters of the gating function.

Now substituting the new value of $\eta_m(x)$ in (12) gives us the new $K_\eta(x_i, x_j)$ which we use to solve the dual formulation (11). We again update the value of the gating function parameters and repeat until convergence. The convergence of the algorithm is determined by observing the change in the objective function in (11). The entire algorithm can be summarized as follows:

**Algorithm 1** *LMKAD* algorithm

1: Initialize the values of $v_m$ and $v_{m0}$ by random values for each $m^{th}$ kernel, where $m = 1, 2, \ldots, p$
2: **do**
3:     Calculate $\eta_m$ using $v_m^t$ and $v_{m0}^t$
4:     Calculate $K_\eta(x_i, x_j)$ using the gating function
5:     Solve conventional one-class SVM with $K_\eta(x_i, x_j)$
6:     $v_m^{(t+1)} \Leftarrow v_m^{(t)} - \mu^{(t)} \frac{\partial J(\eta)}{\partial v_m} \quad \forall m$
7:     $v_{m0}^{(t+1)} \Leftarrow v_{m0}^{(t)} - \mu^{(t)} \frac{\partial J(\eta)}{\partial v_{m0}} \quad \forall m$
8: **while** Not converged

Once the algorithm converges and the final $\eta_m(x)$ and the Lagrange multipliers are obtained, the decision function

can be rewritten as

$$f(x) = \sum_{i=1}^{N} \sum_{m=1}^{p} \alpha_i \eta_m(x) K_m(x, x_i) \eta_m(x_i) - \rho \qquad (13)$$

The sign of this decision function tells us whether the given input is target or outlier. Also, we compute the average error using the sum over the (target value - predicted value) on the training data. Then we set the bias term to this mean value.

In the above discussion, the use case of gating function is explained in detail. Three types of gating functions are used in this paper for our experiments, which are defined as follows:

### 4.1. Softmax Function

$$\eta_m(x) = \frac{exp(\langle v_m, x \rangle + v_{m0})}{\sum_{k=1}^{p} exp(\langle v_k, x \rangle + v_{k0})} \qquad (14)$$

This function is characterized by the parameters $v_{m0}$ and $v_m$. The above function is called the Softmax function and ensures that $\eta_m(x)$ is non-negative. Note that if we use a constant gating function, the algorithm reduces to that of *MKAD* [26], and assigns fixed weights over the entire input space. As per the above discussion, gradient of the objective function $J(\eta)$ needs to be calculated with respect to the parameters of the Softmax gating function ($v_{m0}$ and $v_m$), which is mentioned as follows:

$$\frac{\partial J(\eta)}{\partial v_{m0}} = -\frac{1}{2} \sum_{i=1}^{N} \sum_{j=1}^{N} \sum_{k=1}^{p} \alpha_i \alpha_j \eta_k(x_i) K_k(x_i, x_j) \eta_k(x_j)(\delta_m^k - \eta_m(x_i) + \delta_m^k - \eta_m(x_j)) \qquad (15)$$

$$\frac{\partial J(\eta)}{\partial v_m} = -\frac{1}{2} \sum_{i=1}^{N} \sum_{j=1}^{N} \sum_{k=1}^{p} \alpha_i \alpha_j \eta_k(x_i) K_k(x_i, x_j) \eta_k(x_j)(x_i(\delta_m^k - \eta_m(x_i)) + x_j(\delta_m^k - \eta_m(x_j))) \qquad (16)$$

$$\text{where,} \quad \delta_m^k = \begin{cases} 1, & \text{if } m = k \\ 0, & \text{otherwise} \end{cases}$$

Once we obtain the updated values of the parameters $v_{m0}$ and $v_m$, we calculate the new value of $\eta_m(x)$ using the gating function.

*4.2. Sigmoid Function*

$$\eta_m(x) = \frac{1}{1 + exp(-\langle v_m, x \rangle - v_{m0})} \tag{17}$$

Again the parameters $v_{m0}$ and $v_m$ characterize the above gating function. The gradients of the objective function $J(\eta)$ with respect to the parameters of the sigmoid gating function ($v_{m0}$ and $v_m$) are:

$$\frac{\partial J(\eta)}{\partial v_{m0}} = -\frac{1}{2} \sum_{i=1}^{N} \sum_{j=1}^{N} \alpha_i \alpha_j \eta_m(x_i) K_m(x_i, x_j) \eta_m(x_j)(1 - \eta_m(x_i) + 1 - \eta_m(x_j)) \tag{18}$$

$$\frac{\partial J(\eta)}{\partial v_m} = -\frac{1}{2} \sum_{i=1}^{N} \sum_{j=1}^{N} \alpha_i \alpha_j \eta_m(x_i) K_m(x_i, x_j) \eta_m(x_j)(x_i(1 - \eta_m(x_i)) + x_j(1 - \eta_m(x_j))) \tag{19}$$

*4.3. Radial Basis Function (RBF)*

$$\eta_m(x) = \frac{e^{-\frac{\|x - \mu_m\|_2^2}{\sigma_m^2}}}{\sum_{k=1}^{p} e^{-\frac{\|x - \mu_k\|_2^2}{\sigma_k^2}}} \tag{20}$$

where $\mu_m$ is the center and $\sigma_m$ gives the spread of the local region. Similar as above, the gradients of the objective function $J(\eta)$ with respect to the parameters of the RBF gating function ($\mu_m$ and $\sigma_m$) are:

$$\frac{\partial J(\eta)}{\partial \mu_m} = -\sum_{i=1}^{N} \sum_{j=1}^{N} \sum_{k=1}^{p} \alpha_i \alpha_j \eta_k(x_i) K_k(x_i, x_j) \eta_k(x_j)((x_i - \mu_m)(\delta_m^k - \eta_m(x_i)) + (x_j - \mu_m)(\delta_m^k - \eta_m(x_j)))/\sigma_m^2 \tag{21}$$

$$\frac{\partial J(\eta)}{\partial \sigma_m} = -\sum_{i=1}^{N} \sum_{j=1}^{N} \sum_{k=1}^{p} \alpha_i \alpha_j \eta_k(x_i) K_k(x_i, x_j) \eta_k(x_j)(\|x_i - \mu_m\|_2^2(\delta_m^k - \eta_m(x_i)) + \|x_j - \mu_m\|_2^2(\delta_m^k - \eta_m(x_j)))/\sigma_m^3 \tag{22}$$

$$\text{where,} \quad \delta_m^k = \begin{cases} 1, & \text{if } m = k \\ 0, & \text{otherwise} \end{cases}$$

## 5. Performance Evaluation

In this section, experiments are conducted to evaluate the performance of the proposed classifier over 25 datasets. These datasets are obtained from University of California Irvine (UCI) repository [31] and mainly belong to two disciplines, i.e., medical and finance. Description of the datasets can be found in Table 1. Many of the datasets are slightly imbalanced. Class imbalance ratio of both of the classes are approximately 1 : 2 in case of 9 datasets viz., Iris, Iono(1), Iono(2), Pima(1), Pima(2), German(1), German(2), Wave(1), Wave(2), and. These datasets were originally generated for the binary or multi-class classification task. For our experiments, we have made it compatible with OCC task in the following ways. If a dataset has two classes then we use each of the classes in the binary dataset alternately as the target class and the remaining one as outlier. If a dataset has more than two classes then we use one of the classes in the dataset as the target class and the remaining ones as representative of the outlier class. In this way, we construct 25 one-class datasets from 13 multi-class datasets.

Table 1: Datasets

| S. No. | Name | #Targets | #Outliers | #Features | #samples |
|---|---|---|---|---|---|
| **Small Size Datasets** | | | | | |
| 1 | Iris | 50 | 100 | 4 | 150 |
| 2 | heart(1) | 160 | 137 | 13 | 297 |
| 3 | heart(2) | 137 | 160 | 13 | 297 |
| 4 | Iono(1) | 225 | 126 | 34 | 351 |
| 5 | Iono(2) | 126 | 225 | 34 | 351 |
| 6 | bupa(1) | 145 | 200 | 6 | 345 |
| 7 | bupa(2) | 200 | 145 | 6 | 345 |
| 8 | Japan(1) | 294 | 357 | 15 | 651 |
| 9 | Japan(2) | 357 | 294 | 15 | 651 |
| 10 | Australia(1) | 307 | 383 | 14 | 690 |
| 11 | Australia(2) | 383 | 307 | 14 | 690 |
| 12 | pima(1) | 500 | 268 | 8 | 768 |
| 13 | pima(2) | 268 | 500 | 8 | 768 |
| **Medium Size Datasets** | | | | | |
| 14 | German(1) | 700 | 300 | 24 | 1000 |
| 15 | German(2) | 300 | 700 | 24 | 1000 |
| 16 | Park(1) | 520 | 520 | 29 | 1040 |
| 17 | Park(2) | 520 | 520 | 29 | 1040 |
| 18 | Space(1) | 1541 | 1566 | 6 | 3107 |
| 19 | Space(2) | 1566 | 1541 | 6 | 3107 |
| 20 | Abalone(1) | 2096 | 2081 | 8 | 4177 |
| 21 | Abalone(2) | 2081 | 2096 | 8 | 4177 |
| 22 | Spam(1) | 1813 | 2788 | 57 | 4601 |
| 23 | Spam(2) | 2788 | 1813 | 57 | 4601 |
| 24 | Wave(1) | 1692 | 3308 | 40 | 5000 |
| 25 | Wave(2) | 3308 | 1692 | 40 | 5000 |

*5.1. Experimental Setup*

All the experiments[1] have been conducted on MATLAB 2016a in Windows 7 (64 bit) environment with 64 GB RAM, 3.00 GHz Intel Xeon processor. For implementing existing One-class classifiers, LIBSVM package [32] is used. For every dataset 5 fold Cross-validation (CV) indices are generated and this procedure is repeated 5 times each time constituting a run. These indices are kept same throughout the experiment for all the classifiers. In 5-fold CV, 4 folds are used for training and 1 fold is used for testing. However, out of the 4-folds used for training, only samples from one of the classes (i.e., target class) are used for training the model. Samples from the other classes are used as validation samples to find optimal parameters. We calculate and report the average Gmean for 5-fold CV over 5 runs. Gmean is defined as follows equation:

$$Gmean = \sqrt{precision * recall} \quad (23)$$

Moreover, we have applied statistical tests in order to analyze the results in a better way. To this end, similar to [33], we compute Friedman Rank (**FRank**)[28], Mean of Gmeans (**MGmean**) over all the datasets, and Percentage of the Maximum Gmean (**PMG**). PMG is defined as follows [33]:

$$PMG = \frac{\sum_{i=1}^{\text{no. of datasets}} \left( \frac{\text{Gmean of classifier for } i^{th} \text{ dataset}}{\text{Maximum Gmean achieved for } i^{th} \text{ dataset}} \times 100 \right)}{\text{Number of datasets}} \quad (24)$$

The same experimental setup is followed for evaluating existing and the proposed one-class classifiers. In our experiments, three commonly used kernels are employed which are defined as follows:

1. Linear kernel (l):

   $K_L(x_i, x_j) = x_i^T x_j$

2. Polynomial kernel (p):

   $K_P(x_i, x_j) = (x_i^T x_j + 1)^q$

3. Gaussian kernel (g):
   $K_G(x_i, x_j) = e^{-\frac{(x_i-x_j)^T(x_i-x_j)}{\sigma^2}}$

The order of the Polynomial kernel is chosen through the parameter $q$ to be 2 or 3. The value of $\sigma$ used in the Gaussian kernel is set to the average of the Square Euclidean distance between all the points in the training data.

---

[1] All presented results in this paper are reproducible. Codes with datasets can be found on my GitHub profile https://github.com/Chandan-IITI after the acceptance of the paper

The Linear kernel has no special parameters. The kernel parameter set that has the highest Gmean on the validation folds is considered to be the best configuration and these parameters are used as input along with the training folds for training the model. The trained model is then evaluated over the test set. Since we use 5-fold CV and repeat the experiment five times, for each dataset, we have twenty-five test set results from which we find and report the average Gmean value with standard deviation and the average percentage of support vectors used. We have also performed z-score normalization on each dataset before training and testing.

For comparing our proposed method from the existing kernel-based methods, we have selected 5 popular existing kernel-based one-class classifiers, which are detailed as follows:

(i) Support Vector Machine (*SVM*) based: *OCSVM* [10], *SVDD* [9], *MKAD* [26]

(ii) *KRR*-based: *KOC* [34]

(iii) Principal Component Analysis (*PCA*) based: *KPCA*[35].

*OCSVM* is implemented using LIBSVM[2] library [32]. *SVDD* is implemented by using DD Toolbox[3] [36].

*5.2. Results and Discussion*

In order to illustrate the significance of the multiple kernel approach, we have performed extensive experiments on various combinations of kernels for the existing (*MKAD*) as well as proposed (*LMKAD*) method. In case of *LMKAD*, the combination of kernel name (in small letter) with gating function name (in capital letter) is mentioned in a bracket with the method name. For rest of the methods, single kernel namecombination of kernel name is mentioned using small letterletters with the method name in the bracket. Many combinations are possible with linear (l), polynomial (p) and gaussian (g) kernel, however, only those two combinations of kernels (gpl and gpp) are presented in the paper which have exhibited better performance. Overall, 6 variants of *LMKAD* are generated, namely, *LMKAD(S_gpl)*, *LMKAD(S_gpp)*, *LMKAD(So_gpl)*, *LMKAD(So_gpp)*, *LMKAD(R_gpl)* and *LMKAD(R_gpp)*.

*5.2.1. Performance comparison*

The Gmean of the 5 kernel-based methods over 25 datasets are provided in Tables 2 to 5. Tables 2 and 3 show the results for small-sized datasets and Tables 4 and 5 show the results for medium-sized datasets. Best Gmean per dataset is displayed in boldface in these Tables. Out of 13 small-sized datasets, proposed method performs better for 11 datasets in term of Gmean. For Bupa(2) dataset, *LMKAD(R_gpp)* yields comparable results to *MKAD(gpl)* and

---
[2]https://www.csie.ntu.edu.tw/~cjlin/libsvm/
[3]https://www.tudelft.nl/ewi/over-de-faculteit/afdelingen/intelligent-systems/pattern-recognition-bioinformatics/pattern-recognition-laboratory/data-and-software/dd-tools/

Table 2: Performance in term of average Gmean±standard deviation (%) over 5-folds and 5 runs for small-sized datasets

| One-class Classifier | Iris | Heart(1) | Heart(2) | Iono(1) | Iono(2) | Bupa(1) | Bupa(2) |
|---|---|---|---|---|---|---|---|
| KPCA(g)[35] | 96.44±0.61 | 70.42±0.22 | 63.5±0.78 | 76.54±0.59 | 57.1±0.23 | 62.91±0.4 | 74.28±0.59 |
| KOC(g)[34] | 92.35±0.87 | 65.03±1.1 | 66.39±0.53 | 92.69±0.23 | 53.4±0.85 | 57.09±1.48 | 68.81±0.99 |
| SVDD(g)[9] | 84.12±2.86 | 72.91±0.55 | 64.9±1.4 | 93.13±0.64 | 44.63±0.51 | 60.64±1.23 | 69.75±0.21 |
| OCSVM(g)[10] | 85.06±2.6 | 72.91±0.55 | 64.9±1.4 | 93.13±0.59 | 44.63±0.51 | 60.64±1.3 | 69.78±0.19 |
| OCSVM(p)[10] | 75.18±3.64 | 43.97±2.07 | 32.26±1.91 | 86.46±0.91 | 21.09±3.2 | 56.9±4.14 | 70.63±0.84 |
| OCSVM(l)[10] | 48.52±9.93 | 59.79±6.1 | 56.76±2.86 | 62.28±1.59 | 48.86±3.2 | 52.71±7.22 | 60.25±1.91 |
| MKAD(gpl)[26] | 57.74±0 | 73.4±0 | **67.91**±0 | 83.12±0.11 | 59.91±0 | 64.83±0 | **76.14**±0 |
| MKAD(gpp)[26] | 57.74±0 | 73.4±0 | **67.91**±0 | 86.2±0.27 | 59.91±0 | 64.87±0.05 | **76.14**±0 |
| LMKAD(S_gpl) | 99.81±0.42 | 73.32±0.23 | 67.68±0.53 | 93.99±0.34 | 60.03±0.35 | 64.97±0.44 | 75.67±0.23 |
| LMKAD(S_gpp) | 99.81±0.42 | **73.6**±0.13 | 67.78±0.37 | **95.08**±0.22 | 59.91±0 | **65.12**±0.49 | 75.68±0.21 |
| LMKAD(So_gpl) | **100**±0 | 73.3±0.04 | 67.2±0.51 | 89.64±0.56 | **60.1**±1.01 | 64.79±0.47 | 75.49±0.33 |
| LMKAD(So_gpp) | 99.19±1.82 | 73.34±0.06 | 67.12±0.38 | 89.49±0.54 | 59.84±0.17 | 64.81±0.58 | 75.67±0.67 |
| LMKAD(R_gpl) | **100**±0 | 73.36±0.07 | **67.91**±0 | 88.68±0.25 | 59.92±0.3 | 64.97±0.47 | 75.8±0.41 |
| LMKAD(R_gpp) | 99.61±0.88 | 73.36±0.07 | **67.91**±0 | 89.04±0.45 | 59.83±0.18 | 65±0.5 | 75.87±0.32 |

Table 3: Performance in term of average Gmean±standard deviation (%) over 5-folds and 5 runs for small-sized datasets

| One-class Classifier | Japan(1) | Japan(2) | Australia(1) | Australia(2) | Pima(1) | Pima(2) |
|---|---|---|---|---|---|---|
| KPCA(g)[35] | 64.09±0.29 | 72.29±0.29 | 63.69±0.29 | 73.06±0.18 | 77.98±0.18 | 57.05±0.4 |
| KOC(g)[34] | 67.33±1.24 | 73.48±1.62 | 65.07±0.68 | 74.21±1.12 | 79.04±0.33 | 54.78±0.21 |
| SVDD(g)[9] | 70.15±0.4 | **76.58**±0.28 | 65.55±0.47 | 76.78±0.22 | 79.21±0.19 | 56.71±0.62 |
| OCSVM(g)[10] | **71.45**±0.38 | 75.78±0.29 | 66.08±0.6 | 76.59±0.44 | 79.18±0.19 | 56.59±0.47 |
| OCSVM(p)[10] | 56.84±2.82 | 58.71±0.52 | 56.47±2.65 | 58.14±1.41 | 75.12±0.5 | 51.02±1.25 |
| OCSVM(l)[10] | 62.21±2.89 | 58.54±4.9 | 56.94±2.16 | 63.32±4.58 | 64.9±3.05 | 44.25±3.11 |
| MKAD(gpl)[26] | 66.99±0.08 | 74.87±0.01 | 66.57±0.14 | 75.55±0.1 | 80.59±0.07 | 59.07±0.01 |
| MKAD(gpp)[26] | 66.95±0.16 | 75.13±0.05 | 66.57±0.14 | 75.8±0.08 | 80.61±0.11 | 58.96±0.11 |
| LMKAD(S_gpl) | 66.96±0.16 | 75.83±0.15 | 66.43±0.57 | 76.48±0.18 | 80.69±0.25 | 58.44±0.26 |
| LMKAD(S_gpp) | 67.31±0.27 | 76.56±0.24 | 66.16±0.25 | **77.13**±0.17 | 80.85±0.2 | **59.75**±0.64 |
| LMKAD(So_gpl) | 67.07±0.15 | 76.19±0.2 | 66.44±0.22 | 76.67±0.15 | 80.84±0.2 | 59.02±0.5 |
| LMKAD(So_gpp) | 67.02±0 | 76.23±0.14 | 66.67±0.08 | 76.66±0.16 | 80.69±0.33 | 59.63±0.62 |
| LMKAD(R_gpl) | 67.09±0.1 | 76.22±0.15 | **66.69**±0.19 | 76.56±0.29 | 80.88±0.18 | 58.6±0.31 |
| LMKAD(R_gpp) | 67.1±0.1 | 76.22±0.15 | 66.64±0.09 | 76.64±0.19 | **80.89**±0.13 | 58.52±0.2 |

$MKAD(gpp)$. In case of 12 medium-sized datasets, proposed method yields better Gmean for 6 datasets and comparable Gmean for 3 datasets. When we analyze the impact of gating function and kernel combination on $LMKAD$ then $LMKAD(S\_gpp)$ emerges as the best classifier. Among all $LMKAD$ variants in Tables 2 to 5, $gpp$ kernel combination

Table 4: Performance in term of average Gmean±standard deviation (%) over 5-folds and 5 runs for medium-sized benchmark datasets

| One-class Classifier | German(1) | German(2) | Park(1) | Park(2) | Space(1) | Space(2) |
|---|---|---|---|---|---|---|
| KPCA(g)[35] | 80.77±0.07 | 49.75±0.28 | 70.2±0.18 | 67.79±0.18 | 67.73±0.07 | 68.7±0.05 |
| KOC(g)[34] | 73.17±0.26 | 53.41±0.3 | 96.15±0.17 | **90.74**±0.56 | **72.92**±0.19 | 70.8±0.25 |
| SVDD(g)[9] | 81.1±0.34 | 52.77±0.82 | 97.07±0.23 | 79.77±0.15 | 71.09±0.75 | 71.06±0.1 |
| OCSVM(g)[10] | 80.34±0.33 | 52.8±0.86 | 95.71±0.2 | 81.82±0.31 | 72.39±0.1 | 70.59±0.05 |
| OCSVM(p)[10] | 69.64±0.47 | 24.9±1.25 | 92.15±0.53 | 78.95±0.32 | 70.97±0.23 | **76.64**±0.21 |
| OCSVM(l)[10] | 69.46±1.1 | 48.2±0.89 | 51.07±2.13 | 60.82±2.54 | 54.74±1.73 | 55.83±2.69 |
| MKAD(gpl)[26] | **83.67**±0 | 54.67±0.14 | 94.27±0.18 | 70.88±0.12 | 70.96±0.03 | 71.18±0.05 |
| MKAD(gpp)[26] | 83.64±0.05 | 54.67±0.14 | 96.72±0.16 | 71.53±0.1 | 71.17±0.05 | 71.27±0.04 |
| LMKAD(S_gpl) | 83.13±0.11 | 54.5±0.14 | **99.46**±0.2 | 82.51±0.52 | 72.5±0.11 | 71.4±0.52 |
| LMKAD(S_gpp) | 82.76±0.2 | 54.27±0.05 | 99.26±0.11 | 85.27±0.31 | 72.58±0.13 | 71.68±0.66 |
| LMKAD(So_gpl) | 83.21±0.16 | **55.09**±0.22 | 94.39±0.21 | 71.74±0.68 | 71.69±0.2 | 71.03±0.18 |
| LMKAD(So_gpp) | 83.24±0.22 | 54.74±0.16 | 94.1±0.39 | 71.67±0.24 | 71.67±0.17 | 71.11±0.19 |
| LMKAD(R_gpl) | 83.37±0.2 | 54.72±0.33 | 93.78±0.32 | 71.28±0.13 | 71.55±0.08 | 71.15±0.16 |
| LMKAD(R_gpp) | 83.25±0.19 | 54.76±0.17 | 94.13±0.1 | 71.32±0.26 | 71.59±0.1 | 71.08±0.07 |

Table 5: Performance in term of average Gmean±standard deviation (%) over 5-folds and 5 runs for medium-sized benchmark datasets

| One-class Classifier | Abalone(1) | Abalone(2) | Spam(1) | Spam(2) | Wave(1) | Wave(2) |
|---|---|---|---|---|---|---|
| KPCA(g)[35] | 68.72±0.08 | 68.36±0.06 | 60.33±0.11 | 71.95±1.57 | 47.39±0.16 | 75.25±0.09 |
| KOC(g)[34] | 73.45±0.18 | 73.39±0.13 | **79.06**±0.36 | **82.42**±0.16 | 58.31±0.65 | 69.65±0.35 |
| SVDD(g)[9] | 71.05±2.56 | 72.82±0.46 | 76.64±1.54 | 81.34±0.84 | 64.99±0.07 | 80.09±0.03 |
| OCSVM(g)[10] | 73.06±0.14 | 73.23±0.09 | 78.41±0.24 | 81.66±0.17 | 65.94±0.1 | 78.84±0.09 |
| OCSVM(p)[10] | 71.91±0.16 | 72.09±0.16 | 69.29±0.58 | 71.04±0.44 | 36.22±2.11 | 79.56±0.32 |
| OCSVM(l)[10] | 56.73±2.91 | 54.52±5.24 | 56.93±2.5 | 62.67±1.7 | 45.83±1.95 | 65.66±0.52 |
| MKAD(gpl)[26] | 73.02±0.04 | 72.97±0.09 | 76.99±0.2 | 80.63±0.1 | 63.37±0.05 | 80.95±0.1 |
| MKAD(gpp)[26] | 72.89±0.08 | 72.89±0.15 | 77.2±0.18 | 81.26±0.06 | 64.6±0.05 | 81.06±0.02 |
| LMKAD(S_gpl) | 72.51±0.11 | 73.93±0.48 | 77.44±0.25 | 80.72±0.12 | 66.37±0.2 | **85.09**±0.19 |
| LMKAD(S_gpp) | **74.55**±0.49 | **74.69**±0.3 | 77.82±0.18 | 81.68±0.1 | **66.81**±0.11 | 83.66±0.05 |
| LMKAD(So_gpl) | 73.56±0.59 | 74.3±0.17 | 77.57±0.16 | 80.84±0.08 | 64.17±0.08 | 81.77±0.22 |
| LMKAD(So_gpp) | 73.61±0.34 | 74.42±0.29 | 77.64±0.2 | 80.78±0.09 | 64.17±0.09 | 82.71±0.58 |
| LMKAD(R_gpl) | 73.47±0.2 | 74.14±0.24 | 76.21±1.56 | 80.69±0.51 | 63.63±0.5 | 81.38±0.08 |
| LMKAD(R_gpp) | 73.28±0.15 | 74.19±0.05 | 76.73±1.25 | 80.43±0.41 | 64.16±0.08 | 81.34±0 |

Table 6: Number of datasets for which maximum Gmean has been achieved by One-class Classifiers (in sorted order)

| Position | One-class Classifiers | #Datasets with Maximum Gmean |
|---|---|---|
| 1 | LMKAD(S_gpp) | 10 |
| 2 | KOC(g) | 4 |
| 2 | LMKAD(R_gpl) | 4 |
| 3 | MKAD(gpl) | 3 |
| 3 | LMKAD(So_gpl) | 3 |
| 4 | MKAD(gpp) | 2 |
| 4 | LMKAD(S_gpl) | 2 |
| 4 | LMKAD(R_gpp) | 2 |
| 5 | SVDD(g) | 1 |
| 5 | OCSVM(g) | 1 |
| 5 | OCSVM(p) | 1 |
| 5 | LMKAD(So_gpp) | 1 |
| 6 | KPCA(g) | 0 |
| 6 | OCSVM(l) | 0 |

Table 7: Average percentage of Support Vectors over 5-folds and 5 runs by one-class classifiers for 25 datasets

| One-class Classifier | OCSVM (g) | OCSVM (p) | OCSVM (l) | MKAD (gpl) | MKAD (gpp) | LMKAD (S_gpl) | LMKAD (S_gpp) | LMKAD (So_gpl) | LMKAD (So_gpp) | LMKAD (R_gpl) | LMKAD (R_gpp) |
|---|---|---|---|---|---|---|---|---|---|---|---|
| Iris(1) | 24.30 | 46.20 | 24.60 | 95.30 | 79.60 | 31.10 | 34.60 | 13.50 | 13.60 | 19.30 | 16.70 |
| Heart(1) | 16.72 | 77.00 | 21.56 | 91.53 | 83.84 | 33.09 | 37.03 | 9.72 | 10.16 | 9.59 | 9.59 |
| Heart(2) | 21.82 | 78.82 | 24.49 | 94.68 | 86.68 | 37.24 | 52.75 | 15.25 | 12.65 | 31.61 | 29.02 |
| Iono(1) | 10.24 | 19.47 | 37.24 | 42.09 | 45.00 | 15.16 | 15.04 | 8.47 | 7.31 | 9.73 | 10.07 |
| Iono(2) | 29.40 | 75.55 | 53.49 | 88.02 | 82.46 | 69.30 | 81.46 | 34.51 | 80.33 | 80.55 | 71.75 |
| Bupa(1) | 14.72 | 25.45 | 13.59 | 73.34 | 58.55 | 16.31 | 21.07 | 7.86 | 8.03 | 8.76 | 8.17 |
| Bupa(2) | 14.53 | 17.65 | 11.45 | 76.95 | 61.73 | 14.58 | 21.45 | 9.83 | 10.70 | 11.63 | 9.60 |
| Japan(1) | 11.97 | 37.56 | 14.27 | 91.97 | 80.46 | 18.16 | 21.82 | 9.02 | 16.93 | 16.27 | 15.13 |
| Japan(2) | 11.05 | 51.99 | 12.82 | 83.66 | 77.96 | 15.57 | 21.29 | 7.38 | 7.37 | 7.25 | 7.25 |
| Australia(1) | 11.19 | 37.62 | 14.04 | 91.86 | 79.74 | 25.18 | 28.25 | 50.93 | 47.93 | 34.68 | 34.80 |
| Australia(2) | 10.57 | 53.12 | 12.55 | 89.16 | 81.79 | 15.78 | 19.24 | 6.63 | 6.67 | 6.44 | 6.49 |
| Pima(1) | 8.71 | 19.00 | 7.81 | 54.89 | 46.54 | 8.32 | 10.05 | 5.97 | 6.65 | 6.85 | 6.52 |
| Pima(2) | 10.22 | 33.81 | 10.52 | 78.62 | 66.38 | 11.75 | 16.90 | 8.00 | 6.79 | 12.23 | 8.34 |
| German(1) | 9.61 | 33.27 | 13.21 | 77.23 | 75.81 | 14.91 | 17.66 | 9.38 | 7.80 | 9.93 | 9.72 |
| German(2) | 12.77 | 84.02 | 23.72 | 90.32 | 80.25 | 25.12 | 32.30 | 7.27 | 10.85 | 25.42 | 15.53 |
| Park(1) | 8.24 | 15.87 | 29.86 | 73.08 | 37.60 | 16.15 | 16.82 | 8.03 | 8.56 | 7.89 | 7.28 |
| Park(2) | 8.22 | 34.81 | 25.63 | 40.80 | 48.62 | 15.17 | 20.51 | 10.67 | 7.76 | 7.98 | 8.11 |
| Space(1) | 5.53 | 5.76 | 6.51 | 41.67 | 26.28 | 10.17 | 12.12 | 5.48 | 5.49 | 5.46 | 5.43 |
| Space(2) | 5.47 | 5.69 | 6.52 | 46.42 | 36.60 | 12.39 | 8.73 | 5.47 | 5.42 | 6.09 | 6.53 |
| Abalone(1) | 5.28 | 5.73 | 6.25 | 13.16 | 9.94 | 5.57 | 9.04 | 10.41 | 6.65 | 5.65 | 5.48 |
| Abalone(2) | 5.40 | 6.17 | 5.95 | 15.16 | 10.55 | 5.90 | 6.28 | 8.34 | 5.13 | 6.45 | 5.09 |
| Spam(1) | 6.30 | 26.21 | 10.69 | 58.29 | 55.74 | 6.00 | 6.77 | 5.36 | 5.34 | 6.62 | 6.08 |
| Spam(2) | 5.88 | 32.11 | 9.13 | 46.39 | 45.08 | 5.79 | 9.54 | 5.26 | 5.24 | 8.01 | 7.12 |
| Wave(1) | 6.38 | 70.60 | 9.11 | 91.13 | 55.66 | 9.08 | 10.88 | 6.94 | 6.94 | 7.70 | 6.94 |
| Wave(2) | 5.70 | 16.55 | 7.00 | 56.01 | 37.59 | 19.24 | 26.33 | 15.25 | 16.87 | 26.01 | 16.75 |

Table 8: Friedman ranking (FRank) and Mean of Gmean (MGmean) of all one-class classifiers in increasing order of the FRank.

| One-class Classifier | FRank | MGmean(%) |
|---|---|---|
| LMKAD(S_gpp) | 2.98 | 75.59 |
| LMKAD(So_gpl) | 5.30 | 74.24 |
| LMKAD(S_gpl) | 5.38 | 75.19 |
| LMKAD(So_gpp) | 5.40 | 74.25 |
| LMKAD(R_gpl) | 5.60 | 74.08 |
| LMKAD(R_gpp) | 5.70 | 74.12 |
| MKAD(gpp) | 6.48 | 72.36 |
| MKAD(gpl) | 7.32 | 72.01 |
| OCSVM(g) | 7.70 | 72.86 |
| SVDD(g) | 7.98 | 72.59 |
| KOC(g) | 8.20 | 72.13 |
| KPCA(g) | 11.52 | 68.25 |
| OCSVM(p) | 11.84 | 62.25 |
| OCSVM(l) | 13.60 | 56.87 |

with Sigmoid gating function (i.e., $LMKAD(S\_gpp)$) exhibits better Gmean for 15 datasets and comparable Gmean for rest of the 10 datasets.

It is obvious that multiple kernel learning based methods need more percentage of support vectors compared to conventional $OCSVM$ as more than one kernel is employed for classification. Impact of localization can be observed from Tables 2 to 5 and Table 7 that $LMKAD$ yields better results compared to $MKAD$ for most of the datasets by using significantly lesser percentage of support vectors. Hence, $LMKAD$ leads to sparser solution compared to $MKAD$. Moreover, $LMKAD$ achieves this performance by combining Gaussian kernel with less complex kernel like polynomial and linear kernel.

Further, Gmean-based two other performance criteria viz., MGmean and PMG are computed for all the classifiers to analyze their performances more closely.

*5.2.2. A closer look at the above obtained Gmean using MGmean and PMG*

The performance of each method over the 25 datasets using the MGmean metric is presented in Table 8 and is plotted in a decreasing order in Fig. 1. Based on the obtained results in Fig. 1, it can be clearly stated that all 6 variants of $LMKAD$ have achieved top six positions among all one-class classifiers as per MGmean criterion. Moreover, $LMKAD(S\_gpp)$ yields best MGmean among existing kernel-based one-class classifiers. It is to be noted from Table 6 that $LMKAD(So\_gpp)$ yields maximum Gmean for only one dataset, which is less compared to $KOC(g)$ and rest of the multiple kernel-based methods. However, $LMKAD(So\_gpp)$ holds the third position in Fig. 1 as per MGmean criterion, i.e., yields better MGmean compared to most of the methods mentioned in Table 6. Similar can be experienced with some other methods also. Hence, In order to further analyze the performance of the competing one-class classifiers, the Percentage of the Maximum Gmean (**PMG**) is calculated as per Eq. (24), similar to [33].

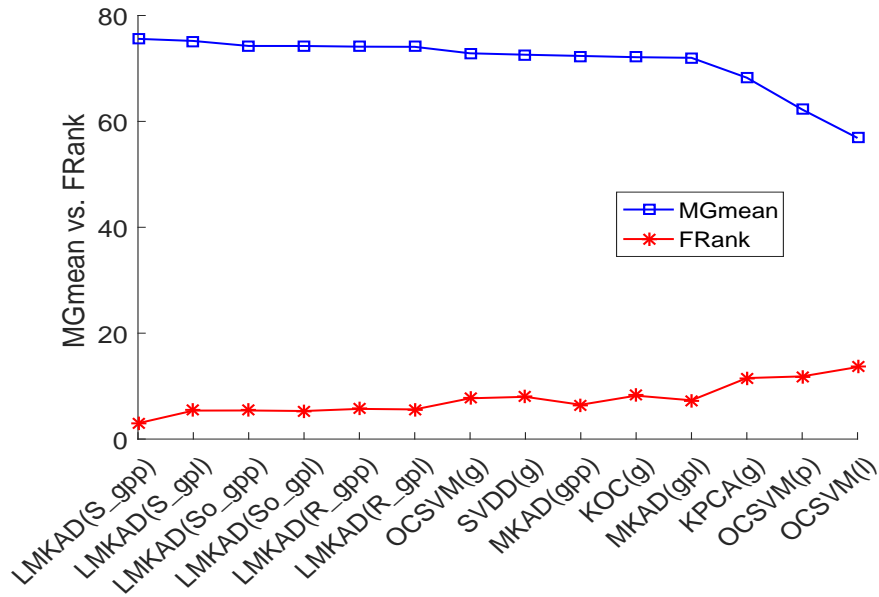

Figure 1: All one-class classifiers as per average Gmean (MGmean) in decreasing order and their corresponding Friedman Rank (FRank).

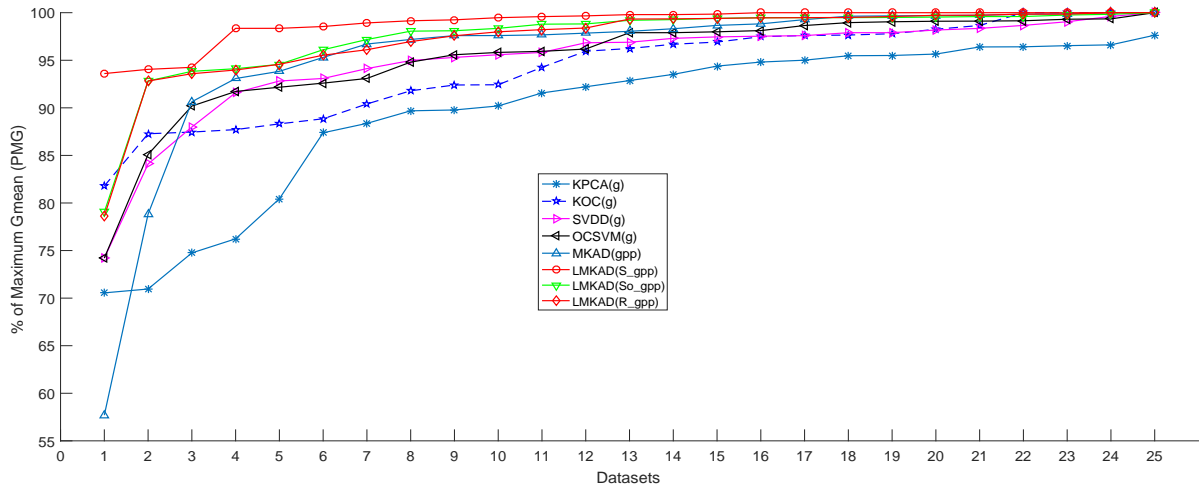

Figure 2: Percentage of the maximum Gmean (PMG) achieved by various one-class classifiers over 15 datasets (ordered by increasing percentage)

PMG metric provides information regarding proximateness of each classifier towards maximum Gmean value. As it can be seen in Table 9, all 6 variants of *LMKAD* hold the top six positions similar to the ranking based on the MGmean values in Fig. 1. In Fig. 2, PMG values of 3 variants of *LMKAD* and 5 existing one-class classifiers are plotted in an increasing order for all 25 datasets. *LMKAD* is a multi-kernel version of the *OCSVM(g)*. The plotted lines for these two classifiers in Fig. 2 clearly indicate the performance improvement of multi-kernel version over single-kernel one. Overall, Fig. 2 illustrates the clear superiority of the localized multi-kernel-based one-class classifiers over the existing methods. Moreover, *LMKAD(S_gpp)* and *LMKAD(S_gpl)* obtain more than 97% PMG value for all datasets except Japan(1), Space(1) and Park(1) datasets. Even for these three datasets, *LMKAD(S_gpp)* and *LMKAD(S_gpl)* achieve more than 90% PMG value. Detailed PMG values for all one-class classifiers over 25 datasets are made available on the web page (https://goo.gl/DuYdJE).

Table 9: Percentage of the Maximum Gmean (PMG)

| One-class Classifiers | PMG (%) |
|---|---|
| LMKAD(S_gpp) | 98.91 |
| LMKAD(S_gpl) | 98.38 |
| LMKAD(So_gpp) | 97.36 |
| LMKAD(So_gpl) | 97.35 |
| LMKAD(R_gpp) | 97.18 |
| LMKAD(R_gpl) | 97.15 |
| MKAD(gpp) | 95.42 |
| OCSVM(g) | 95.28 |
| MKAD(gpl) | 95.02 |
| SVDD(g) | 94.91 |
| KOC(g) | 94.24 |
| KPCA(g) | 89.72 |
| OCSVM(p) | 80.13 |
| OCSVM(l) | 75.41 |

Above discussion suggests *LMKAD(S_gpp)* as the best performing classifier in term of Gmean, MGmean, and PMG. Despite this fact, a statistical testing needs to perform for verifying this fact. In the next subsection, Friedman Rank (FRank) testing is performed for statistical testing.

*5.3. Statistical comparison*

For comparing the performance of the proposed and existing kernel-based methods on 25 benchmark datasets, a non-parametric Friedman test is employed. In the Friedman test, the null hypothesis states that the mean of individual experimental treatment is not significantly different from the aggregate mean across all treatments and the alternate hypothesis states the other way around. Friedman test computes three components viz., F-score, p-value and Friedman Rank (FRank). If the computed F-score is greater than the critical value at the tolerance level $\alpha = 0.05$, then one rejects the equality of mean hypothesis (i.e. null hypothesis). We employ the modified Friedman test [28] for the testing, which was proposed by Iman and Davenport [37]. The F-score obtained after employing non-parametric Friedman test is 24.56, which is greater than the critical value at the tolerance level $\alpha = 0.05$ i.e. $24.56 > 1.75$. Hence, we can null hypothesis can be rejected with 95% of a confidence level. The computed p-value of the Friedman test is $2.1454e - 28$ with the tolerance value $\alpha = 0.05$, which is much lower than 0.05. This small value indicates that differences in the performance of the various methods are statistically significant.

Afterwards, FRank of each classifier is also calculated to assign a rank to all presented one-class classifiers in this paper. Friedman test assigns a rank to all the methods for each dataset, it assigns rank 1 to the best performing algorithm, the second best rank 2 and so on. If rank ties then average ranks are assigned [28]. The FRank of all one-class classifiers is provided in increasing order in Table 8 and is visualized in Fig. 1 with decreasing order of MGmean. $LMKAD(S\_gpp)$ still achieves top position, similar to using the MGmean metric. From Table 8 and Fig. 1, it can be observed that FRank of most of the classifiers follows a similar pattern as MGmean, i.e., FRank increases as MGmean decreases. However, some of the one-class classifiers don't follow the same pattern as with MGmean. For example, $LMKAD(S\_gpl)$ has better MGmean but inferior FRank compared to $LMKAD(So\_gpl)$. The above analysis indicates that a one-class classifier with better FRank has better generalization capability compared to the other existing methods.

Overall, after the performance analysis of all one-class classifiers, it is observed that none of the existing one-class classifiers perform better than the proposed *LMKAD* one-class classifier in term of MGmean, PMG, and FRank.

## 6. Conclusion

In this paper, an anomaly detection/OCC method (*LMKAD*) is proposed as an extension of *MKAD*. *LMKAD* provides a localized formulation for multi-kernel learning method by local assignment of weights to each kernel. The derived formulation is also shown to be analogous to conventional *OCSVM* and is solved in a similar fashion using a LIBSVM solver. In *LMKAD*, local assignments of weights are achieved by the training of a gating function and a conventional *OCSVM* with the combined kernel in tandem. The proposed method is empirically tested with 3 types of gating function and over 25 benchmark datasets. Performance of *LMKAD* is compared against 5 kernel-based one-class classifiers. *LMKAD* outperforms both conventional *OCSVM* and *MKAD* for most of the datasets. For some other datasets, it performs similar to *MKAD* but uses fewer support vectors and provides a sparser solution. The Friedman test is also performed, which verifies that the experimental outcomes are statistically significant. In future work, proposed localized formulation can be extended for other variants of kernel-bases one-class classifier.

**Acknowledgment**

This research was supported by Department of Electronics and Information Technology (DeITY, Govt. of India) under Visvesvaraya PhD scheme for electronics & IT.